\documentclass[a4paper,conference]{IEEEtran}
\IEEEoverridecommandlockouts
\usepackage{cite}
\usepackage{amsmath,amssymb,amsfonts}
\usepackage{algorithmic}
\usepackage{graphicx}
\usepackage{textcomp}
\usepackage{xcolor}
\usepackage{float}
\def\BibTeX{{\rm B\kern-.05em{\sc i\kern-.025em b}\kern-.08em
 T\kern-.1667em\lower.7ex\hbox{E}\kern-.125emX}}
\begin{document}

\title{Sparse Image based Navigation Architecture to Mitigate the need of precise Localization in Mobile Robots
}

\author{\IEEEauthorblockN{Pranay Mathur}
\IEEEauthorblockA{\textit{Mobile Robotics} \\
\textit{Addverb Technologies}\\
Noida, India \\
pranay.mathur@addverb.com}
\and
\IEEEauthorblockN{Rajesh Kumar}
\IEEEauthorblockA{\textit{Mobile Robotics} \\
\textit{Addverb Technologies}\\
Noida, India \\
rajesh.kumar01@addverb.com }
\and
\IEEEauthorblockN{Sarthak Upadhayay}
\IEEEauthorblockA{\textit{Mobile Robotics} \\
\textit{Addverb Technologies}\\
Noida, India \\
sarthak.upadhyay@addverb.com }

}

\maketitle

\begin{abstract}
Traditional simultaneous localization and mapping (SLAM) methods focus on improvement in the robot's localization under environment and sensor uncertainty. This paper, however, focuses on mitigating the need for exact localization of a mobile robot to pursue autonomous navigation using a sparse set of images. The proposed method consists of a model architecture - RoomNet, for unsupervised learning resulting in a coarse identification of the environment and a separate local navigation policy for local identification and navigation. The former learns and predicts the scene based on the short term image sequences seen by the robot along with the transition image scenarios using long term image sequences. The latter uses sparse image matching to characterise the similarity of frames achieved vis-a-vis the frames viewed by the robot during the mapping and training stage. A sparse graph of the image sequence is created which is then used to carry out robust navigation purely on the basis of visual goals. The proposed approach is evaluated on two robots in a test environment and demonstrates the ability to navigate in dynamic environments where landmarks are obscured and classical localization methods fail.
\end{abstract}


\section{Introduction}
\begin{figure*}[htp]
 \centering
 \includegraphics[width=\textwidth,height=5cm]{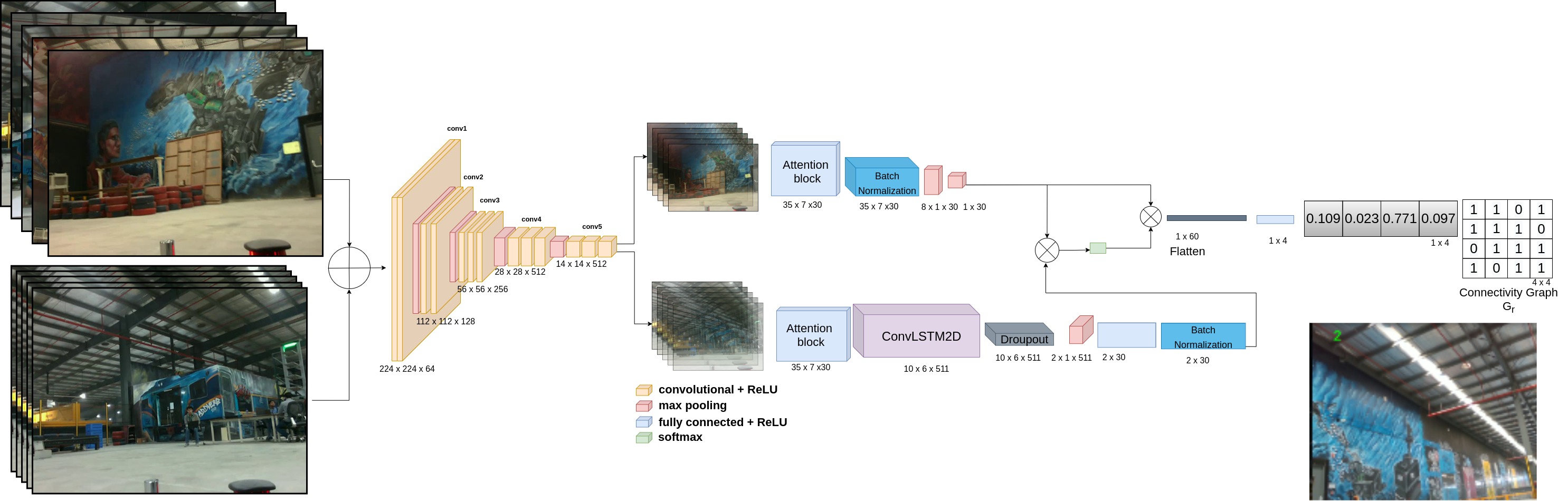}
 \caption{RoomNet Network Architecture: The input consists of a queue $Q_{I_t}$ of current images and a queue $Q_{I_h}$ of images at a fixed interval. The network output is a tensor which gives us the room ID $r$}
 \label{Fig:Network}
\end{figure*}

Simultaneous Localization and Mapping (SLAM) based methods have been popular since the 2000s. The methods focus on localizing the robot within an environment while simultaneously creating the map of the environment. Traditional SLAM methods are known to perform remarkably well on high quality data \cite{fuentes2015visual, durrant2006simultaneous, bailey2006simultaneous}. However, the methods are prone to errors in situations where the sensor data is noisy or when the environment itself is non-static leading to errors. The SLAM methods focus on the odometry data attached to the robot which is prone to drift and then rely on matching the environment to the map in order to prescribe the exact location within the environment. The localisation is usually described in terms of the Euclidean coordinates described in $\mathbb{R}^2$, requiring an exact prescription of a globally fixed zero frame. Intermediate localisation errors lead to incorrect actions by the robot. Numerous methods have been proposed in the literature in order to improve the robot's localization in the presence of inaccurate sensor input, changing environment, change in the location of the robot within the environment and others. Existence of repetitive features within an environment is another issue which leads to inaccurate loop closure and hence, inaccurate localizations. Outdoor localisation has been aided by the existence of Global Positioning Systems (GPS) but the indoor localisation has been dependent on the on board sensors on the robot \cite{cadena2016past, chen2020survey}.
\par Recently, different methods have been proposed to pursue robot navigation in different conditions. This includes robust odometry estimation techniques including the utility of visual odometry \cite{nister2004visual, forster2014svo, yang2020d3vo}, visual-inertial odometry \cite{li2013high, zuniga2020vi} and LiDAR (Light Detection and Ranging) based odometry techniques \cite{zhang2014loam}. Apart from traditional LiDAR based localization, image based localization techniques are popular \cite{zhang2006image}.
\par Apart from the classical models based on fundamentals from vision, deep learning methods are very popular recently as they tend to capture non linear behaviour which is difficult to be modelled using classical strategies. Visual Odometry techniques like DeepVO \cite{wang2017deepvo} have focussed on utilising the recurrent convolutional neural networks leveraging the memory saving feature of the same. Structure from motion (Sfm) synthesis techniques have also been developed  \cite{zhou2017unsupervised} for visual odometry.
\par In order to develop a robust SLAM pipeline, different methods of scene representation are popular. Dense scenes are reconstructed by fusing the RGB and the depth images \cite{kerl2013dense}. Monocular image based understanding of the scene provides a compact descriptor of the scene itself \cite{wang2021tt, lee2021improved}. Other representations of the scene like the point cloud representation \cite{newcombe2011dtam}, voxel representation \cite{ji2017surfacenet}, mesh representation \cite{young2021sparse}, and semantic representation \cite{bao2021utilization, gonzalez2021s3lam} have been used. Again, the neural network based SLAM methods including the CNN-SLAM \cite{tateno2017cnn},  CodeSLAM \cite{bloesch2018codeslam}, SceneCode \cite{zhi2019scenecode}, FASTSLAM \cite{montemerlo2002fastslam}, SLAM-net \cite{karkus2021differentiable} are able to train a robust end-to-end SLAM architecture. CodeSLAM and SceneCode focuses on training a reduced version of the scene whereas CNN-SLAM is an end-to-end predictor. SLAM-net architecture proposes a differentiable architecture where the end-to-end SLAM architecture is fused into a differentiable computation graph, ensuring backpropagation of the loss for each of the sub-module of the traditional SLAM pipeline.
\par Many of the methods focus on improvement on the existing SLAM and mapping based navigation pipelines using neural network based methods. It is well known that most of the industrial use case for mobile robots using SLAM pipelines have utility in warehouse scenarios \cite{zou2021comparative}. The challenging and changing industrial environments have been well documented in the literature and challenging datasets like the \textit{Scene Change Detection} dataset are made available to tackle \cite{park2021changesim}. Due to the existence of featureless walls and rapidly changing scenarios, exact localization based methods fail. The fail cases are usually related to the occasions of moving features, existence of featureless structures and rapidly changing methods. Some of the real life scenarios where the autonomous mobile robot, Dynamo, failed to localize based on the existing map has been shown in Figure \ref{fig:Evaluation} . It can be seen that in a true industrial scenario many of the features are dynamic and repeated exact localization of the robot in the environment is difficult. However, there are certain elements like the pillar or the exact room exits are stationary and do not change throughout. The researchers have been trying to mitigate the issue by incorporating the methods \cite{rosen2016towards, liu2021lifelong}. There has been an attempt to relate the semantically understood feature components within the scene \cite{wang2021structured}. Infact, visual navigation language (VLN) models have been increasingly popular \cite{wang2018look, wang2019reinforced}.
\par This paper focuses on mitigating the case of exact localization of the robot within the environment and characterizing the understanding of the robot's location in terms of entry-exit image scenarios for the region. A sparse image based graph consists of the entry exit image the robot needs to see while navigating the environment. Additionally, the goal to the robot is an image scene rather than an exact coordinate to the robot. The method reduces the dependency of exact localization of the robot in terms of the Euclidean coordinates. The main contributions of our work titled RoomNet can be quantified as:
\begin{itemize}
	\item Describing a neural network leveraging the short term and the long term scenic memory of the robot to locate itself with respect to a room ID and trained on videos stored while passively navigating
	\item Describing an image based local navigation policy where the robot repeatedly aims for a sequence of goal images in order to achieve the desired location.
\end{itemize}

\section{Network}
\begin{figure*}
 \centering
 \includegraphics[width=\textwidth,height=6cm]{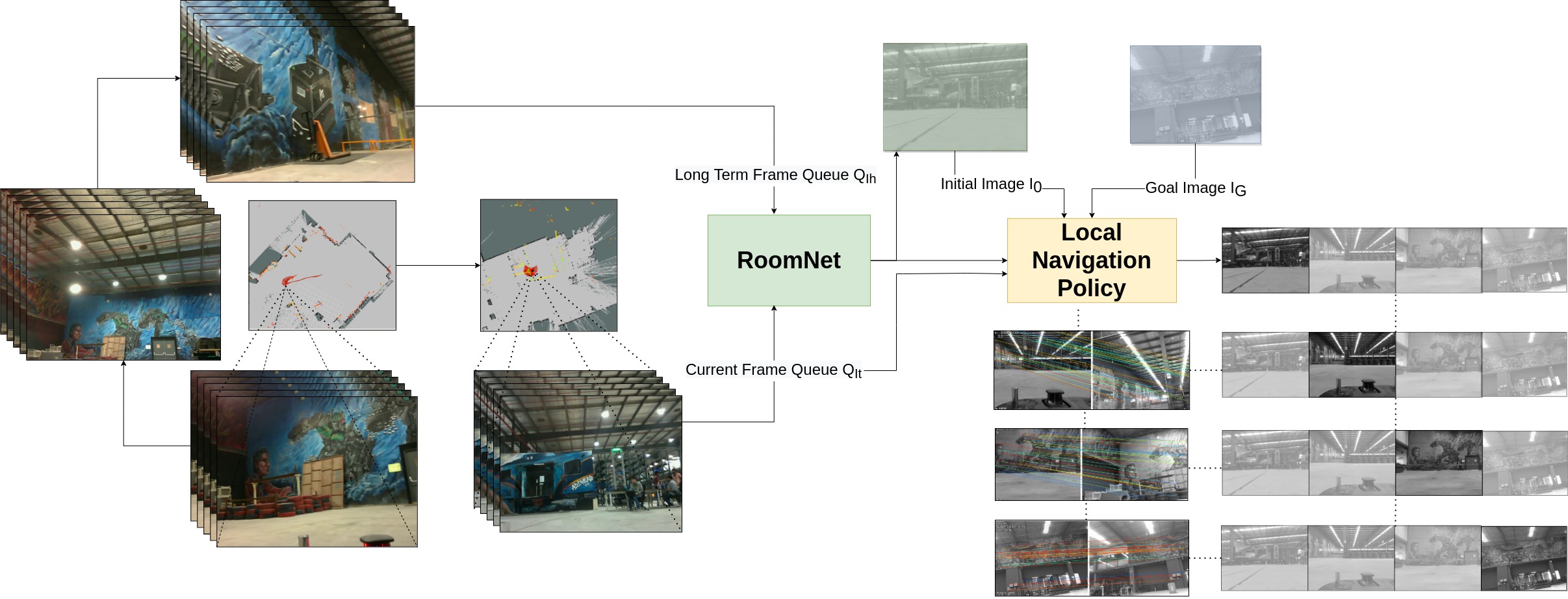}
 \caption{Illustration of our method}\label{method}
\end{figure*}
The paper focuses on describing a network so that the robot is able to understand a sparse location in terms of the image scenes needed to cross to exit the scene. Throughout the paper, we call the scene a room. The room can be either characterised by a physical room entity or a hypothetical scenario associated with a discrete change in the scene. A R-CNN based neural network is created based on long term and short term memory. The short term memory is characterised by a LSTM (Long Short Term Memory) architecture whereas the long term memory is defined from a memory bank.
A set of pretrained VGG-16 backbone is used to characterise the features that the robot sees. The pretrained backbone architecture results in quick feature generation and less number of trainable parameters which can be trained during the initial mapping stage.
\par Let the scene be discretized into `m' rooms, each of them with two transition locations. Each of the transition locations is characterised using a set of transition frames that the robot sees (coined as `transit' in the paper). The short term memory bank constitutes a sequence of `$n_1$' images, sampled at the time interval of $t_1$. The long term memory bank constitutes a set of `$n_2$' images, sampled at the time intervals of $t_2$ with $t_2 >> t_1$. All the $n_1 + n_2$ frames sequentially pass through the VGG-16 feature extractor backbone. The concatenated set of short term memory features pass through the LSTM block. However, the long term features are sparsely selected over a long range of time and might not capture the exact understanding of the room for longer instances. In order to mitigate the issue, an attention model is described to capture the long term memory of the robot.
\par \textbf{VGG-16}: VGG16 is a CNN based model architecture which was proposed during 2014  \cite{szegedy2015going}. A representation of the backbone is presented in the Figure \ref{Fig:Network}. The network uses small $3\times3$ sized filters with a fixed stride of 1, thereby capturing small discrete features within the image. The padding is used such that the spatial resolution is affine to the convolution operation. The model, although has a lot of trainable parameters has been widely used for the operations of object detection and classification. In this work, the feature backbone is used to create an ensemble of features to be trained.
\par \textbf{LSTM}: LSTM or Long-Short Term Memory is a sequence based recurrent neural network \cite{hochreiter1997lstm}. A representative view of the LSTM structure is shown in the Figure \ref{Fig:Network}. The LSTM network consists of a forget gate, input gate and the output gate. The short term memory features of the local scene seen by the robot is mapped to the room ID and thence, to the transition sequence of the frames.
\par \textbf{Attention Block:} Attention models have shown promise in context of complex sequence modelling where the elements of sequence are not sampled at a consistent frequency \cite{vaswani2017attention}. It was assumed that the robot's local velocity is constant, thereby leading to approximate constant feature sampling for the short term memory component of the frames. However, the long term feature vectors when sampled based on a constant frame frame, cannot be at a constant frequency because the robot itself does not move at a globally constant velocity. The attention block is presented in the Figure \ref{Fig:Network}. The context features are described from the saved frames in the long term memory whereas the current input feature is based on the current image frame.
\par \textbf{Complete Model:} The network model utilises a combination of the instantaneous current frame, a set of short term sequence of frames recently seen by the robot and a set of long term frames seen by the robot. The context features for each of the frames is described by the feature detector backbone (VGG-16) and then bifurcated to an LSTM and an attention block. The two outputs are concatenated to a final softmax layer which presents the probability of the robot being indexed with a set of entry - exit sequence of images.

\section{Method}
The proposed method divides the problem into two sub-modules. A network architecture - RoomNet for coarse identification of the scene and a separate module for navigation and finer recognition using a local navigation policy.

\par \textbf{RoomNet}: RoomNet operates in two stages in an unsupervised manner. A training and mapping stage during which a predefined set of maneuvers is executed and an inference stage for navigation.  
During the training and mapping stage, a sparse graph $G_r(V,E)$ is created with the room ID number $r \in \mathbb{W}$, as vertices being mapped to the scene viewed by the robot. From the queue of image frames, $Q_{I} $ associated with $r$, an image sub-queue, $Q_{I_t}$, is chosen by selecting frames post a fixed time interval. These frames are used to train the model along with the older images, $Q_{I_h}$ serving the purpose of a long-term memory.  They take into account the transition images associated with an edge $E$ of $G_r(V,E)$ and the frames the robot could have seen prior to the current scene. The sparse graph formulates the connections between different nodes thereby providing additional constraints on the output inference of RoomNet.

\par Once the sparse graph is created, the goal to the robot is fed in terms of a goal image, $I_g$. The goal image associates with it a set of frames based on the environment that it is located in. These are the frames the robot ``needs to view'' to associate it with a room ID $r$ where the scene is known to be available. During initialization, the network RoomNet is fed the queue, $Q_{I}$,  consisting of the latest image frames. This provides a location that is used to initialize the additional module. It associates the room ID with an initial frame $I_o$ acting as the source node in the graph. Simultaneously, the user-defined goal image $I_g$, is passed. Using $Dijkstra's$ algorithm\cite{dijkstra1959note}  a hierarchy is then generated from the sparse graph for the room ID's, $r \in (0,1,...m)$ to visit and the transition images, $I_{r}$ the robot needs to look for. Thus, the output of RoomNet is a $1\times(m+1)$ tensor with an associated probability $p_m$ corresponding to the prediction. The tensor is multiplied with $G_r(V,E)$ to get the final output matrix.

\begin{equation}
G_r =
\begin{bmatrix}
r_{00} & r_{01} & r_{02} & ... & r_{0m} \\
r_{10} & r_{11} & r_{12} & ... & r_{1m} \\
\vdots & \vdots  & \vdots & ... & \vdots \\
r_{m0} & r_{m1} & r_{m2} & ... & r_{mm} \\
\end{bmatrix}
\end{equation}

\par \textbf{Local Navigation Policy}: Once a frame has been associated with a scene the policy is used for local navigation. Images are matched using graph neural-network based feature-matching \cite{sarlin2019superglue}. Intermediate images are said to have been identified when the matching score $m_s$ calculated as shown in (\ref{Eqn:matching_score}) crosses a pre-defined threshold $m_{s0}$. The intermediate goal image $I_r$, is said to have been reached when the gradient $v$, of scores generated based on matching the current frame to the intermediate goal image $I_{k}$ turns negative and the product of the instantaneous matching scores and uncertainty measure of inference from RoomNet $C_t$, over a moving window of $l$ previous predictions, drops below a threshold $C_o$.

\begin{equation}
m_s = \frac{Features \; Matched}{Keypoints\; Detected \;in\; Query\; Image}
\label{Eqn:matching_score}
\end{equation}

\begin{equation}
v = \eta \nabla m_s
\end{equation}

\begin{equation}
C_{t} = \sum_{i=0}^{l} \frac{1}{e^{{|r-r_{t-i}|}}}
\end{equation}
\begin{figure}[H]
 \includegraphics[width=\columnwidth]{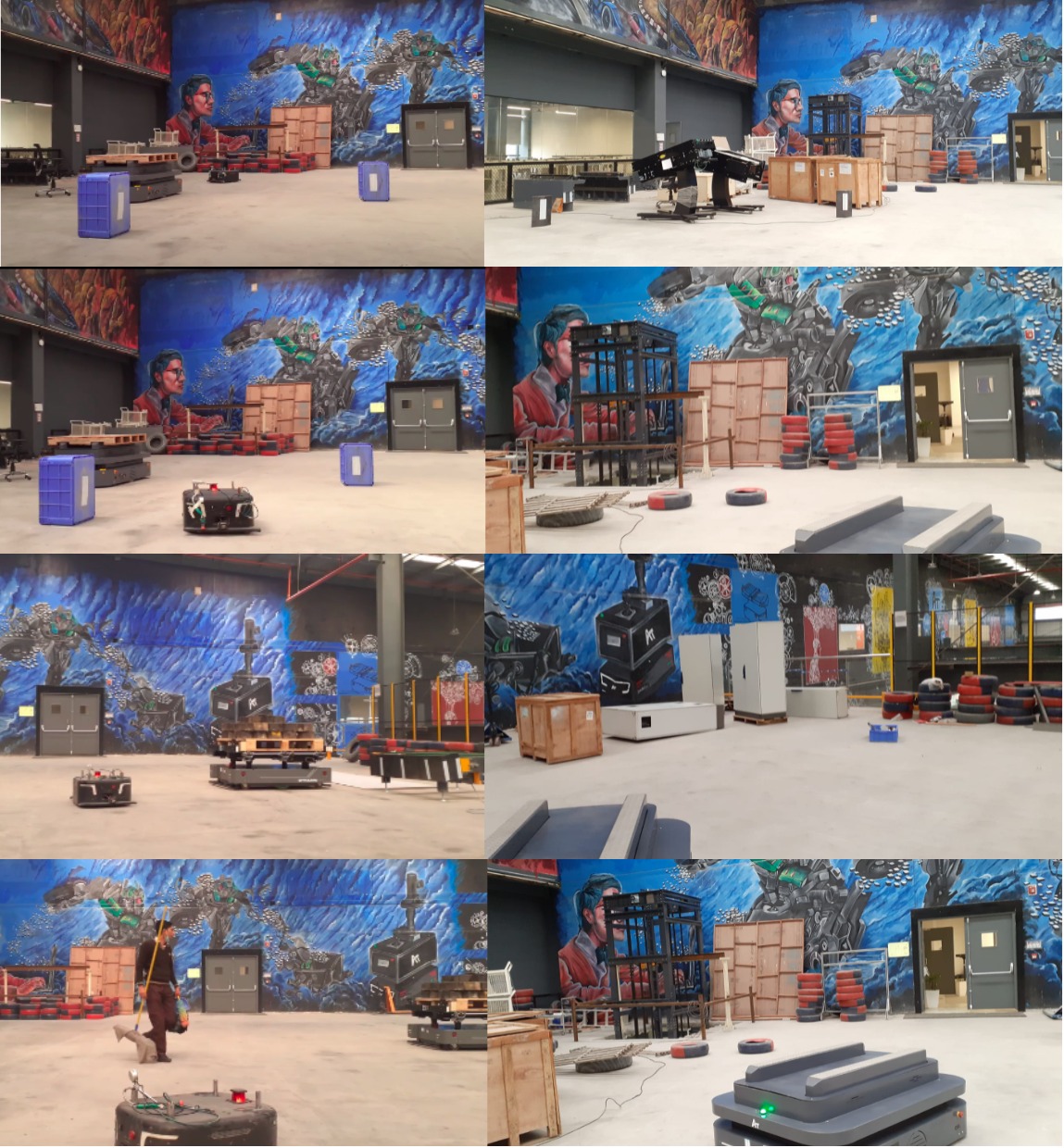}
 \caption{Test Environment: The left column presents the original scene while the right column visualizes the same scene with new obstacles and altered configurations. The test robots Dynamo 100 and Dynamo 1500 are visible in the last row on the left and right column respectively}
 \label{fig:scene_change}
\end{figure}

\begin{figure*}[!htp]
 \includegraphics[width=\textwidth,height=21cm]{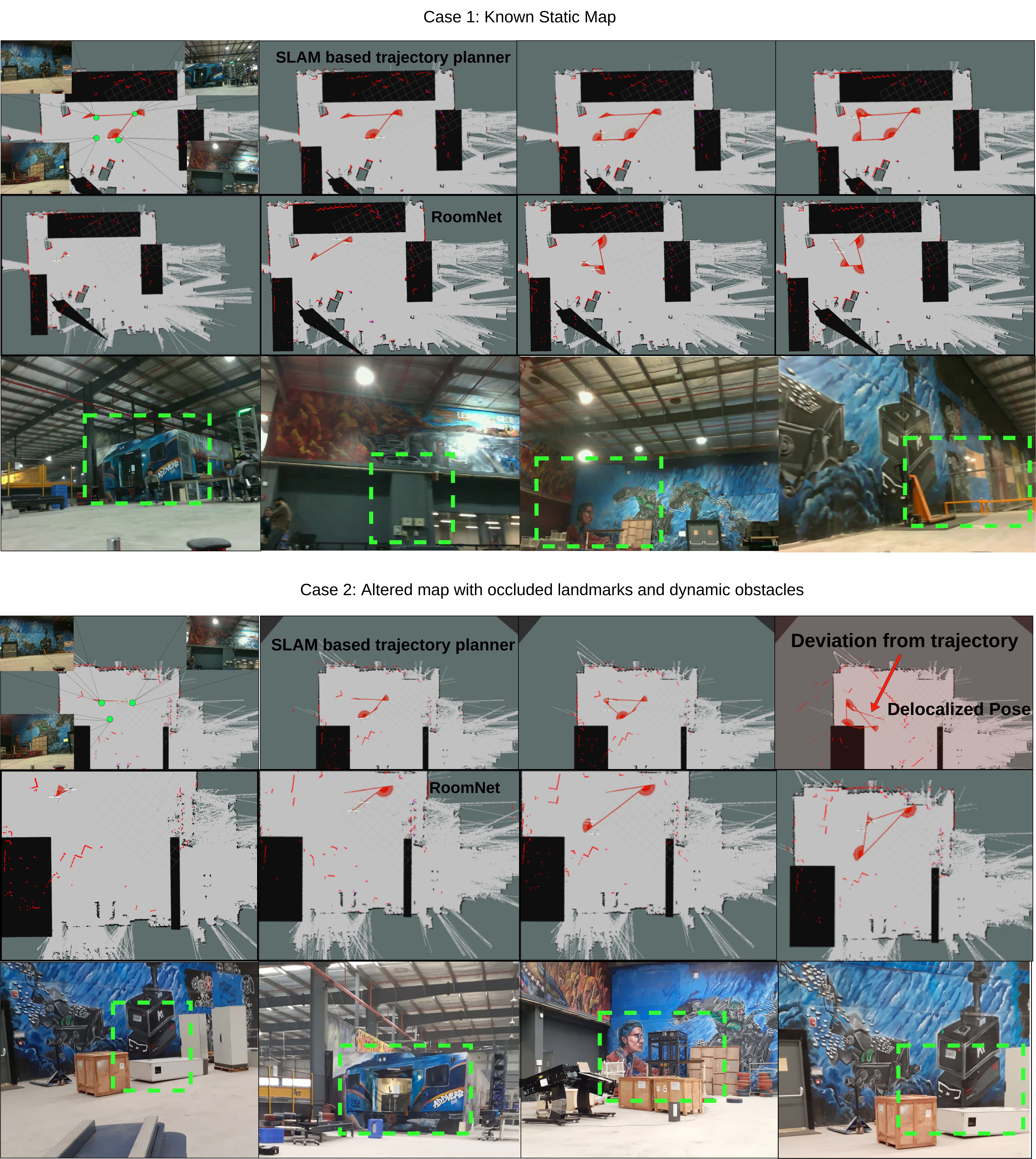}
 \caption{Evaluation: The red arrows indicate the trajectory of the robot. In case 1, The goal points highlighted in green are placed in the shape of a trapezium. In case 2, the goal points highlighted in green are placed to form the vertices of a triangle. The red points are visualized laser scans}
 \label{fig:Evaluation}
\end{figure*}
\subsection{Implementation}
Our approach has been implemented on two robots - a Dynamo 100 and Dynamo-1500\footnote{https://addverb.com/product/autonomous-mobile-robot} robot shown in Figure \ref{fig:scene_change}, each equipped with a 2.3 GHz Intel celeron processor, SICK LiDARs\footnote{https://www.sick.com/in/en/detection-and-ranging-solutions} and an Intel Realsense D435i depth camera\footnote{https://www.intelrealsense.com/depth-camera-d435i}. The industrial safety grade LiDAR is used purely for obstacle avoidance and plays no role in our algorithm.
Training of RoomNet is performed using videos collected while passively navigating. The scenes are annotated with a room ID according to the the scene they belong to.
Upon initialization we perform in-place rotation for RoomNet to provide an initial inference. Post this an image associated with the inference is identified as the first target image. Upon identification of the target image the robot moves along the shortest path possible while avoiding obstacles as described in the section above. When the target is reached a new image with the next room ID, $r$ is associated. This process repeats till the final goal is reached. In case the robot enters a room that is out of the hierarchical sequence of room ID's, $Dijkstra's$ algorithm\cite{dijkstra1959note} is run again taking this node as the source and the goal node as the destination.

\section{Results and Evaluation}
We evaluate our results in a scene that consists of  4 rooms. An exemplary set of scene change images is presented in the Figure \ref{fig:scene_change}. The desired image goal target associated with our local navigation policy is highlighted in dotted green rectangles. We perform evaluation by comparing our method with the GMapping SLAM\cite{grisetti2007gmapping} and adaptive Monte Carlo localization\cite{fox1999amcl} based trajectory planner in a dynamic environment.
\par Consider the first case where the robot is localized as is made evident by the laser scans in red coinciding with the obstacle boundaries. The traditional SLAM architecture based trajectory planner plans a quadrilateral shaped path with the goal points highlighted in green. The trajectory followed by the robot is presented in the Figure  \ref{fig:Evaluation}. The result of our approach is demonstrated in the next row of images. The  local planner followed the path more accurately and the time taken was lower in comparison.\par Consider the second case, where the room had changes and dynamic obstacles. Here we observe the traditional SLAM architecture led to an error in localization as is evident by the altered laser scans. The trajectory followed by the robot is presented in the Figure \ref{fig:Evaluation} and the trajectory post delocalization have been highlighted in red. Our approach was able to follow the planned triangle shaped trajectory accurately and reach the final goal. For the purpose of visualization of the trajectory we manually initialized the pose upon delocalization. The image goals were reached successfully. We reiterate that the map is used purely for visualization purposes and our approach does not require a metric map.
A video of our results can be found at https://youtu.be/KwP2NuIrdx4

\section{Conclusion}
This work focuses on mitigating the need of exact localization of the mobile robot in terms of exact euclidean coordinates and focusing on the local short term and the long term scenes that the robot sees. A neural network is described to associate a set of transition images with the current scene viewed by the robot. The robot then directs itself towards repetitive transition views and ultimately to the goal. The use of long term and short term memory scenes have helped sustain consistency even when the environment suffered dynamic changes.
\bibliographystyle{ieeetrans}
\bibliography{sparse_image.bib}
\end{document}